\documentclass[12pt]{article}
\usepackage{wrapfig}
\usepackage{graphicx} 

\usepackage[dvipsnames]{xcolor}
\usepackage[most]{tcolorbox}
\usepackage{amssymb}
\usepackage{booktabs}
\usepackage{tabularx}
\usepackage{caption}
\usepackage{subcaption}

\usepackage[sorting=none]{biblatex}
\usepackage{mdframed}
\usepackage{hyperref}
\usepackage{array} 

\usepackage[a4paper, left=0.8in, right=0.8in, top=1in, bottom=1in]{geometry}
\usepackage{setspace}
\begin{document}

\title{%
\includegraphics[width=0.5\linewidth]{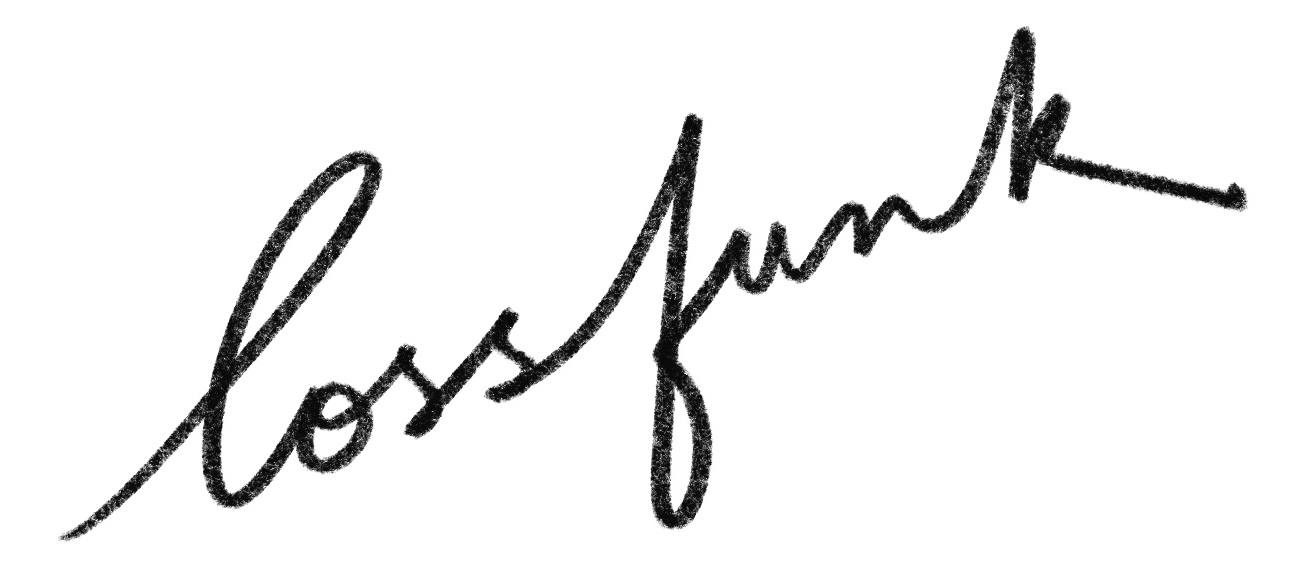} 
\\[1cm]
Why LLMs Aren't Scientists Yet: Lessons from Four Autonomous Research Attempts
}

\author{
  Dhruv Trehan \hspace{2cm} Paras Chopra \\
  \small \texttt{\{dhruv.trehan, paras\}@lossfunk.com}
}
\date{January 2026}

\maketitle

\begin{abstract}
We report a case study of four end-to-end attempts to autonomously generate ML research papers using a pipeline of six LLM agents mapped to stages of the scientific workflow. Of these four, three attempts failed during implementation or evaluation. One completed the pipeline and was accepted to Agents4Science 2025, an experimental inaugural venue that required AI systems as first authors, passing both human and multi-AI review. From these attempts, we document six recurring failure modes: bias toward training data defaults, implementation drift under execution pressure, memory and context degradation across long-horizon tasks, overexcitement that declares success despite obvious failures, insufficient domain intelligence, and weak scientific taste in experimental design. We conclude by discussing four design principles for more robust AI-scientist systems, implications for autonomous scientific discovery, and we release all prompts, artifacts, and outputs at \url{https://github.com/Lossfunk/ai-scientist-artefacts-v1}.
\end{abstract}



\setlength{\parskip}{1em}   
\setlength{\parindent}{0pt} 

\section{Problem Definition and System Overview} \label{sec:systemoverview}

The question we set out with was: could state-of-the-art reasoning LLMs go from a research idea to a research paper with a high degree of autonomy, minimal code scaffolding, and the most basic tools? Various AI Scientist systems proposed in research papers already relied on a high degree of domain-specific pre-definition of the workflow or framing the problem statement in a system-specific way. Tree-search systems like Sakana's \cite{yamada2025aiscientistv2workshoplevelautomated} would have required complex meta-orchestration, contradicting our minimal scaffolding goal. Similarly, Google's AlphaEvolve system \cite{novikov2025alphaevolvecodingagentscientific} requires a clear verification metric defined by a human expert in advance. 

\begin{figure}[h!] 
        \centering 
        \includegraphics[width=1\textwidth]{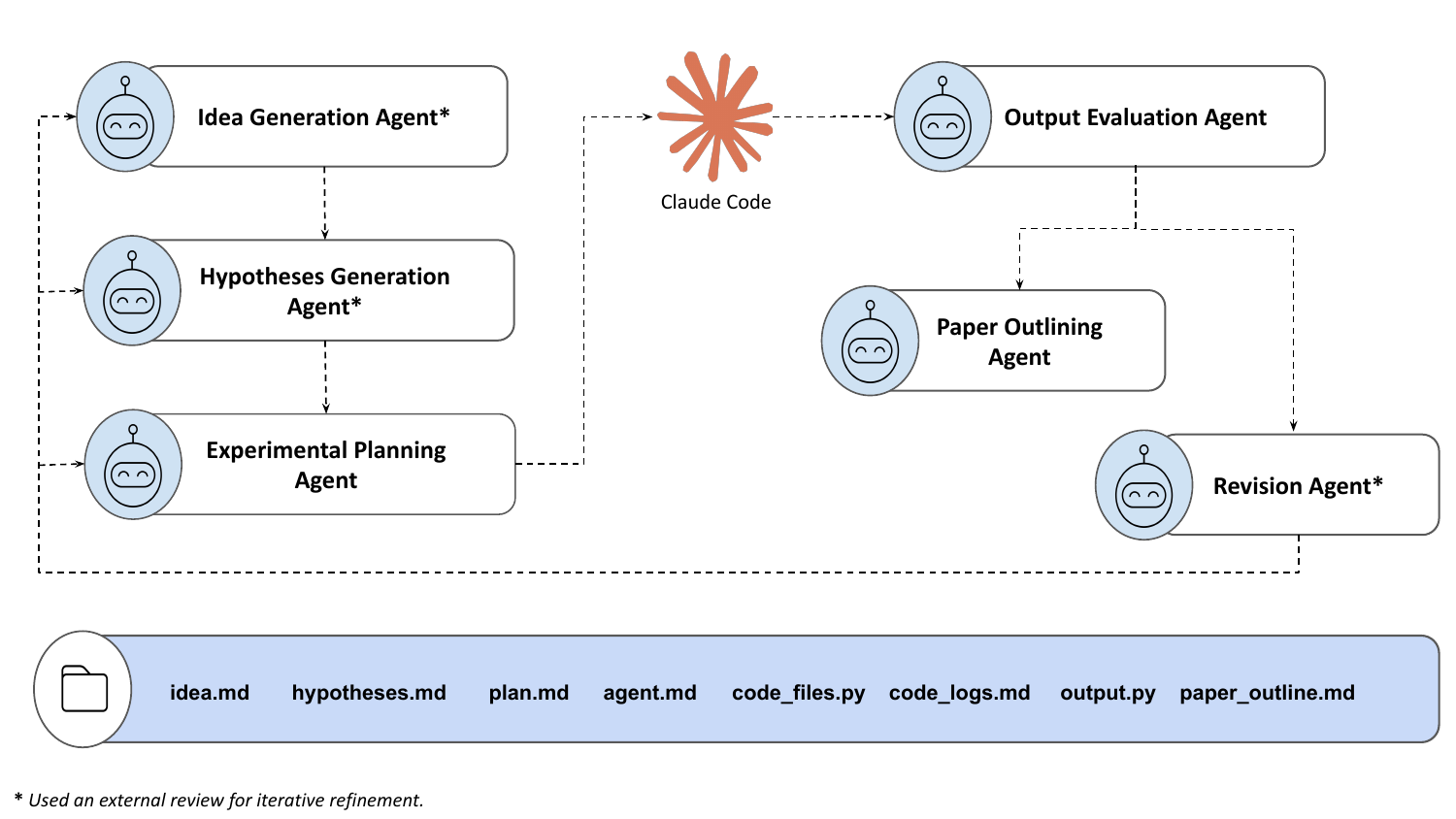} 
        \captionsetup{justification=centering, labelfont=bf}
        \caption{\textbf{Autonomous Research Pipeline}: High-level diagram showing the interaction between the six agent modules and the shared file system artifacts (\texttt{idea.md} to \texttt{paper\_outline.md}) used to maintain context.} 
        \label{fig:system_design} 
\end{figure}

We were interested in exploring how far current LLMs can go without significant scaffolding or inputs from humans. In line with this, we limited our scope to computational sciences, and specifically Machine Learning, since experiments in this domain could be done completely digitally. Figure~\ref{fig:system_design} shows a high-level diagram of our system design.

Our final system consisted of the following six distinct modules, all using Gemini 2.5 Pro primarily because of its long context length. Prompts and output formats for each of these are available in our GitHub repository\footnote{\url{https://github.com/Lossfunk/ai-scientist-artefacts-v1}}.

\begin{enumerate}
      \item \textbf{Idea Generation Agent}, which generated research ideas by combining insights from two input papers in a subdomain and outputted an idea in a structured format. More details on the idea generation process are included in Section~\ref{sec:attemptsoutcomes}.

      \item \textbf{Hypotheses Generation Agent}, which took the generated research idea and proposed testable and falsifiable hypotheses with clear claim statements, datasets, baselines, and metrics to observe to guide experimental planning in subsequent stages.

      \item \textbf{Experiments Planning Agent}, which converts hypotheses and ideas into detailed implementation plans (\texttt{plan.md} and \texttt{agent.md} files) containing step-by-step tasks, project structure, method clarifications, and failure mode controls for autonomous execution by Claude Code on Modal infrastructure. 

      \item \textbf{Experimental Output Evaluation Agent}, which employed a two-tiered evaluation approach, primarily reviewing the output of an experiment for hypothesis implementation fidelity and statistical validity. The same agent was later used to conduct a paper readiness check for a completed hypotheses suite to determine if sufficient insight had emerged to proceed to paper writing tasks.
      
      \item \textbf{Revision Agent}, which automatically determined the next step in the research process when experimental output evaluation indicated failure, choosing between revising the idea, revising the hypotheses suite, or requesting a LLM-mentor feedback. The agent was triggered infrequently since most failed experiments were terminated by human decision rather than routed through the revision process.
      
      \item \textbf{Paper Outlining Agent}, which reviewed the complete experimental output, as well as all relevant context docs across stages, to outline a research paper including descriptions of visualizations required. Claude Code then completed this outline, section by section, into a full paper.
\end{enumerate}

\textbf{Experimental Implementation and Paper Writing} -
The experimental implementation and paper writing were executed by Claude Code using Claude Opus 4.1 and Claude Sonnet 4 for iterative code development, execution on Modal infrastructure, and manuscript development from outline to full text. 

For paper-writing, we adopted a three-stage, iterative approach: generating a detailed outline, completing sections sequentially, and conducting two rounds of human-Claude Code collaborative editing—first to ensure continuity and flow, then to temper overoptimistic claims about results. In our experience, human intervention remained necessary for quality control during the paper-writing phase because initial autonomous drafts generated with prompting were technically accurate but tended toward overly mechanical prose or lacked narrative nuance.

\textbf{Tools and Model Access} -
The model was accessed on Google AI Studio using an agentic prompt, including repository location and the following four tool definitions:
  \begin{itemize}
      \item \texttt{read\_file}  - This enabled the LLM to read context files including previously generated documents, mentor notes from external LLM review, or outputs from code implementation and execution. 
      \item \texttt{write\_file} - This enabled the LLM to save context files. 
      \item \texttt{llm\_search} - This enabled agents to query current information from the internet via an OpenAI model when needed throughout the research process.
      \item \texttt{list\_files} - This returned a list of files in the context repo. This becomes more relevant as the project progresses and the LLM agent creates sub-directories. 
  \end{itemize}
  
\setlength{\columnsep}{25pt}
\begin{wrapfigure}{r}{0.5\textwidth}   
  \centering
  \vspace{-30pt}
  \includegraphics[width=0.5\textwidth]
  {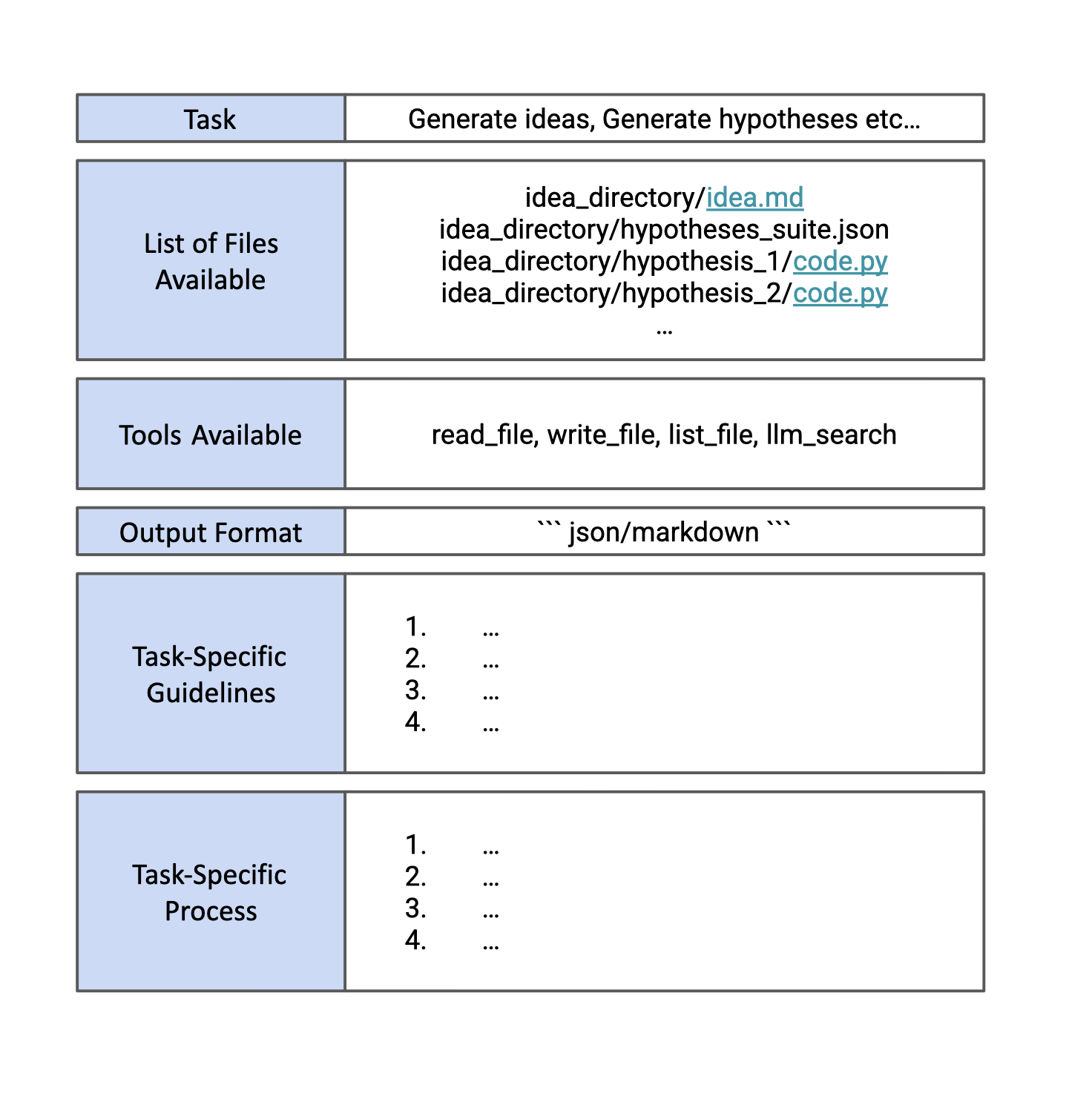}
  \vspace{-20pt}
  \captionsetup{justification=centering, labelfont=bf}
  \caption{\textbf{Agent Prompt Template}: How repository state, tools, and process guidelines are shared in the system prompt of each agent module.}
  \label{fig:promptoutline}
  \vspace{-10pt}
\end{wrapfigure}

\textbf{Context Management} -
Each research idea operated within a dedicated repository that served as the workspace for all agents and the context store. This repository included all context artefacts created from idea note, to hypotheses sets, code files, and code outputs, as well as the final paper LaTeX template required for the conference submission.

Each agent received the repository state as part of its prompt context, as shown in Figure~\ref{fig:promptoutline}. This approach ensured that we were not context engineering too much, in line with our autonomy and general system design constraints, and we could review the LLM's decision-making about which context files to parse when, a relevant information retrieval skill for any researcher.

\textbf{Idea Review} -
Along with this, we also used four zero-shot prompts (NeurIPS Guidelines Based Reviewer \cite{yamada2025aiscientistv2workshoplevelautomated}, Chain of Ideas Reviewer \cite{li2024chain}, Google Co-scientist Tournament Reviewer \cite{gottweis2025aicoscientist}, and a custom evaluation prompt with web-search access via OpenAI) for idea review and shortlisting before the hypotheses generation and experimentation process began. All reviewer prompts were run with Claude Opus 4.1, and only the custom evaluation prompt had web-search access. \footnote{We decided to not include web-search with 3/4 reviewer prompts in line with conversations about the human paper review process in which reviewers mentioned that they did not conduct much literature review before writing reviews.}

\section{Our Attempts and Outcomes} \label{sec:attemptsoutcomes}

\begin{wrapfigure}{r}{0.45\textwidth} 
    \centering
    \vspace{-15pt} 
    \includegraphics[width=\linewidth]{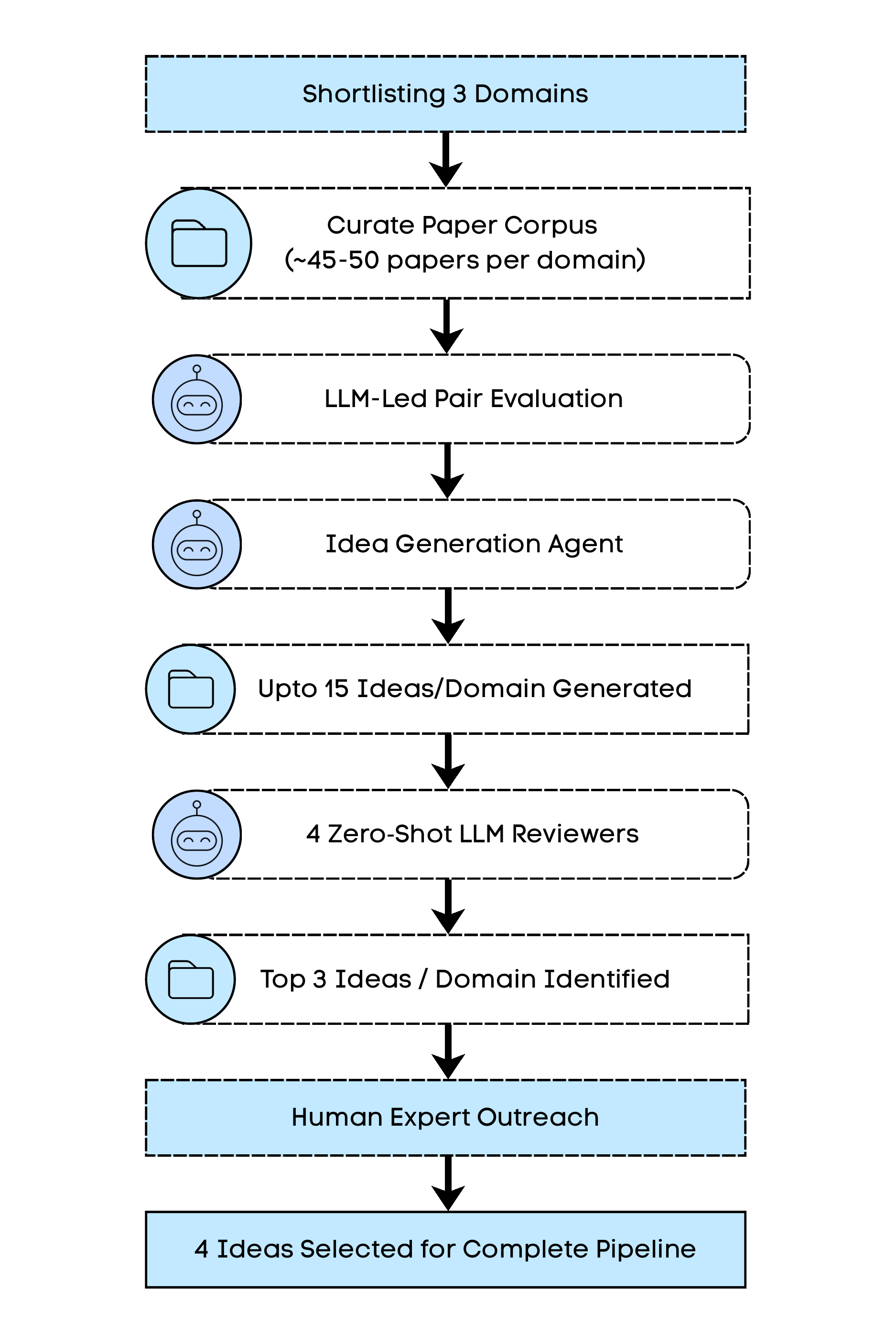} 
    \captionsetup{justification=centering, labelfont=bf}
    \vspace{-5pt}
    \caption{\textbf{The Selection Funnel}: From 135+ papers to 4 candidates. Only one (AS-1) survived execution constraints.}
    \label{fig:attempts_flowchart}
    \vspace{-30pt}
\end{wrapfigure}

To test our autonomous research system, we explored three domains: World Models, Multi-Agent Reinforcement Learning, and AI Safety and Alignment. Figure~\ref{fig:attempts_flowchart} shows our multi-stage process for generating and reviewing ideas, leading to the final four ideas that progressed from idea generation to hypothesis generation and experimentation. We started with a corpus of papers from top-tier venues to ensure our idea generation was grounded in cutting-edge research, and the final decision on which ideas to pursue was made after contacting the authors of the parent papers for their input on quality and feasibility.

\setlength{\columnsep}{25pt} 

Of these four ideas, three were pruned during early execution, and only the fourth (from the AI Alignment and Safety domain) was successfully completed. Table \ref{tab:research_implementation_summary} shows a high-level summary of these four candidate ideas, with final status and primary failure modes. For clarity, we refer to each research idea by its ID (MARL-1, WM-1, WM-2, AS-1) throughout the remainder of this document. Detailed case studies for each attempt, including their complete text, source papers, expert validation, and specific errors, are provided in Appendix~\ref{sec:deepdives}. 

\begin{table}
    \centering
    \captionsetup{justification=centering, labelfont=bf}
    \caption{Research Implementation Journey: Attempts, Outcomes, and Failure Modes}
    \label{tab:research_implementation_summary}
    \vspace{0.5em}
    \scriptsize 
    \renewcommand{\arraystretch}{1.3} 
    
    
    \begin{tabularx}{\textwidth}{@{} l l
    >{\raggedright\arraybackslash}X c
    >{\raggedright\arraybackslash}X @{}
    >{\raggedright\arraybackslash}X @{}}
    
    \toprule
    \textbf{ID} & \textbf{Domain} & \textbf{Idea} & \textbf{Runs} & \textbf{Final Status} & \textbf{Primary Failure Modes} \\
    \midrule
    
    \textbf{MARL-1} & Multi-Agent RL & Zero-shot Multi-agent Coordination & 2 & 
    Failed: Execution Stage & 
    Implementation Drift, Bias on Training Data \\
    \addlinespace
    
    \textbf{WM-1} & World Models & Differentiable Planning in Stochastic World Models & 1 & 
    Failed: Evaluation Stage & 
    Implementation Drift, Lack of Scientific Taste \\
    \addlinespace
    
    \textbf{WM-2} & World Models & Replacing Reconstruction Loss with Perceptual Loss & 1 & 
    Failed: Evaluation Stage & 
    Implementation Drift, Bias on Training Data \\
    \addlinespace
    
    \textbf{AS-1} & AI Safety & Using Semantic Entropy for Jailbreak Detection & 2 & 
    Success: Published at Agents4Science 2025 & 
    Bias on Training Data, Memory and Context Issues \\
    
    \addlinespace
    
    \bottomrule
    \end{tabularx}
\end{table}

Results from the fourth idea that was successfully executed became our final paper, accepted to the Agents4Science 2025 conference: ``The Consistency Confound: Why Stronger Alignment Can Break Black-Box Jailbreak Detection.'' \textbf{The conference accepted 48/254 valid submissions, and our paper also passed a code-audit instituted by the conference organizers.} 

Reviewers in the conference included three distinct LLM Reviewers and a Human Expert. Table \ref{tab:semantic_entropy_reviews} shows the reviews that primarily recognized the work for its empirical contribution and negative results, while also noting that the analysis and comparisons were limited, leading to a majorly \textbf{borderline accept} decision, with 1 of 3 AI reviewers scoring the paper 6 or \textbf{strong accept}.

\begin{table}[h!]
\centering
\captionsetup{justification=centering, labelfont=bf}
\caption{Agent4Science 2025 Reviews for ``The Consistency Confound''}
\label{tab:semantic_entropy_reviews}
\vspace{0.5em}
\scriptsize
\renewcommand{\arraystretch}{1.3}
\begin{tabularx}{\textwidth}{@{} l c c >{\raggedright\arraybackslash}X >{\raggedright\arraybackslash}X @{}}
\toprule
\textbf{Reviewer Type} & \textbf{Decision} & \textbf{Confidence} & \textbf{Strengths} & \textbf{Weaknesses} \\
\midrule
        AI Reviewer 3 & 4 & 5 & Solid contribution by revealing fundamental limitations of a plausible detection approach and providing mechanistic understanding of why it fails. & The focus on a single SE variant and limited model families. \\
\addlinespace
        AI Reviewer 2 & 6 & 5 & The paper's originality lies not in the proposal of a new method, but in its rigorous deconstruction of an existing one in a new context. & No significant weaknesses identified. \\
\addlinespace
        AI Reviewer 1 & 4 & 5 & Clear empirical negative results with careful quantification, rigorous analysis of failure modes, appropriate baselines, and transparent discussion of limitations and ethical considerations. & Limited scope (the SE variant differs from canonical SE), lack of comparison to content-based black-box detectors, calibration protocol concerns, narrow decoding and embedding choices, and no positive alternative proposed. \\
\addlinespace
        Human Reviewer & 4 & 3 & The paper is technically sound, with careful empirical evaluation. It identifies a clear and reproducible failure mode. The underlying method is borrowed from hallucination detection. Main originality is in showing its limitations rather than developing a new technique. & Scope is limited, comparisons are mostly against simple baselines, with no testing of stronger black-box defenses. Contribution is primarily a negative result. While important, significance would be higher if the paper also explored alternative methods or mitigation strategies. \\
\bottomrule
\end{tabularx}
\end{table}

As part of the conference submission, we had to include an AI Involvement Checklist to measure and explain the role of AI in our research. Table~\ref{tab:ai_involvement} shows a snippet of this. We rated the system at the highest autonomy level (Category D, $\ge$95\% AI) for experimental design, execution, and writing, with human involvement (Category C, 50-95\% AI) primarily in the initial hypothesis definition stage. The complete paper, code generated, specific agent logs, and the full checklist are available for review on OpenReview \cite{trehan2025the}.

\begin{table}[h!]
    \centering
    \captionsetup{justification=centering, labelfont=bf}
    \caption{AI Involvement Checklist for Idea AS-1 by Research Stage (Agents4Science Rubric)}
    \label{tab:ai_involvement}
    \vspace{0.5em}
    \scriptsize
    \renewcommand{\arraystretch}{1.4}
    \begin{tabularx}{\textwidth}{@{} l c X @{}}
    \toprule
    \textbf{Research Stage} & \textbf{Score} & \textbf{Role Description} \\
    \midrule
    Hypothesis Development & \textbf{C} & \textit{Mostly AI}: Initial search space defined by human; specific failure mode hypothesis and core questions generated by AI agents via paper mashing. \\
    \addlinespace
    Experimental Design & \textbf{D} & \textit{AI-Generated}: Full experimental plans, baseline choices, and metric definitions generated by Gemini 2.5 Pro; human role limited to high-level approval. \\
    \addlinespace
    Execution \& Analysis & \textbf{D} & \textit{AI-Generated}: End-to-end coding by Claude Code. Datasets handled autonomously. Human intervention limited to providing HuggingFace tokens. \\
    \addlinespace
    Paper Writing & \textbf{D} & \textit{AI-Generated}: Narrative structuring and figure generation driven by AI. Human role restricted to final ``sanity check'' and minor copy-editing. \\
    \bottomrule
    \addlinespace[1em]
    \multicolumn{3}{l}{\footnotesize \textbf{Score Key:} A: $\ge$95\% Human, B: 50--95\% Human, C: 50--95\% AI, D: $\ge$95\% AI.}
    \end{tabularx}
\end{table}

\section{Observed Failure Modes and Mitigation} \label{sec:failuremodes}

Through our experiments, we identified six critical failure modes that consistently emerged across different research attempts. These patterns reveal systematic limitations in current LLM capabilities for autonomous research. The following subsections detail each failure mode with specific examples and mitigation strategies.


\subsection{Bias on Training Data}

\textbf{Research often relies on specialized protocols, libraries, and datasets that aren't widely used, but models consistently defaulted to popular alternatives from their training data.} This failure mode manifested systematically across infrastructure setup, library selection, and dataset handling, where models consistently overrode explicit instructions with presumably memorized training patterns. 

\setlength{\columnsep}{25pt} 
\begin{wrapfigure}{r}{0.5\textwidth} 
    \centering
    \captionsetup{justification=centering, labelfont=bf}
    \vspace{-20pt} 
    \includegraphics[width=\linewidth]{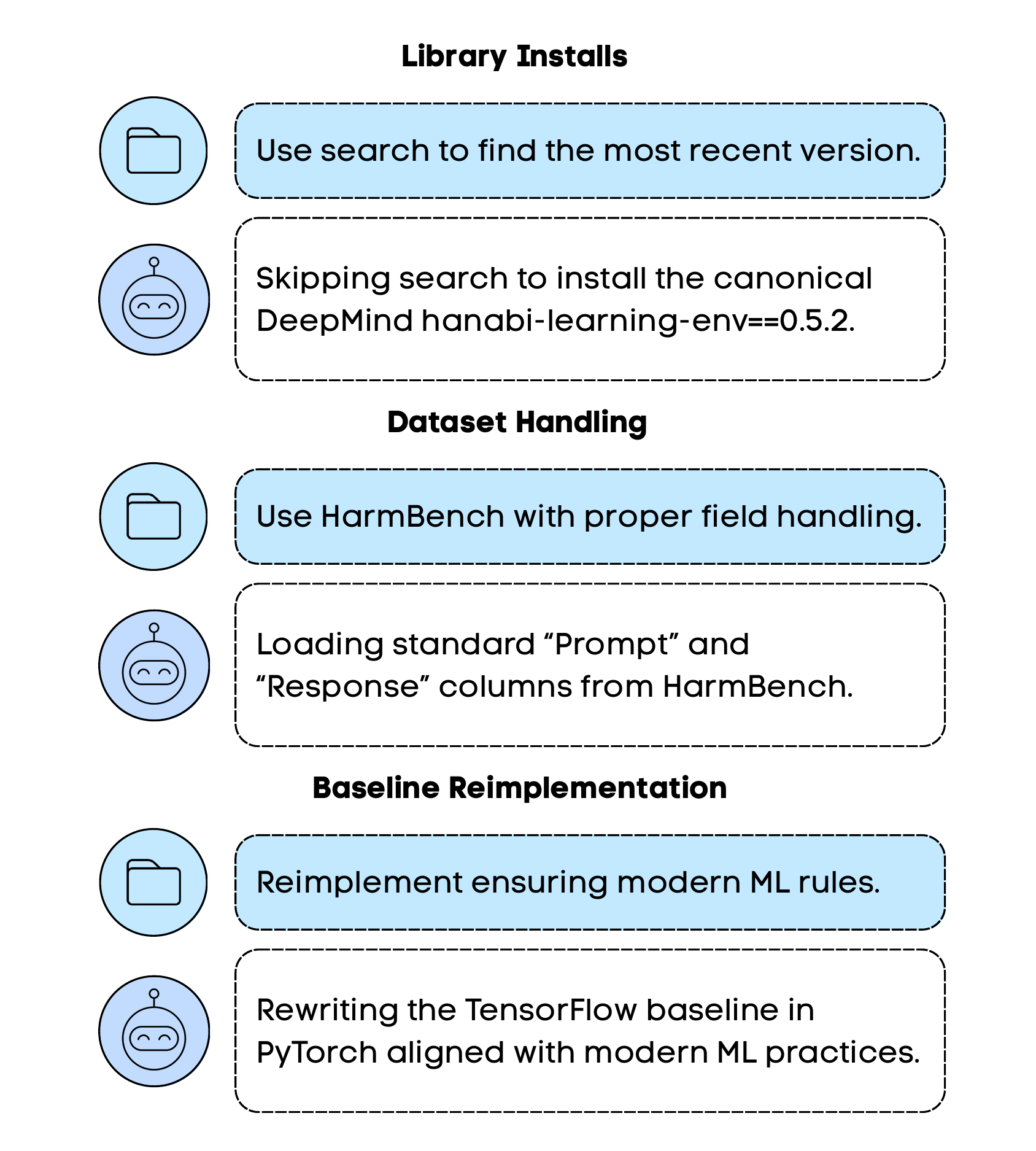}
    \caption{\textbf{Training Data Bias Pattern}: Systematic override of prompt instructions by memorized patterns across long contexts.}
    \label{fig:training_data_bias}
    \vspace{-10pt} 
\end{wrapfigure}

In the case of Modal setup, Claude Code would consistently use an outdated Modal \texttt{mount} command to make local files accessible, and default to local file paths over Modal storage, ignoring updated API documentation and instructions provided in \texttt{plan.md} and \texttt{agent.md} file. Beyond this, it would also regress to outdated research libraries despite explicit version and search requirements, and ignore dataset-specific field structures in favor of standard formats. Figure~\ref{fig:training_data_bias} showcases this systematic override across three implementation contexts, with detailed examples provided in Appendix~\ref{sec:deepdives} per idea implementation.

Similar errors in setting up libraries and environments were found by researchers working specifically on this problem \cite{arora2025setupbenchassessingsoftwareengineering} \cite{eliseeva2025envbenchbenchmarkautomatedenvironment}. \textbf{We addressed this by avoiding anchoring on low-level details like libraries to use and datasets at earlier stages, and providing instructions for specific library use and library documentation at the execution stage.} 

Even after these two changes, in the case of an error, the experiment execution model would often diagnose library imports as the cause and regress back to using the versions mentioned in the training data, insisting that that is the right way to do it.

This bias in training data can show up in more nefarious ways than just errors in using libraries and datasets. Mehtaab Sawhney and Mark Sellke observe with regard to their work on the Erdős problems with GPT-5 Pro, ``while models are able to suggest plausible proof strategies, they often...are overly confident in the power of existing methods.'' They note that this is unsurprising, since discussion of why ``a more obvious strategy'' doesn't work is ``largely absent in mathematical literature itself'' and, consequently, in the training data. In this case, the bias on training data leads to a failure of domain taste and identifying what they call ``negative space'' for a problem \cite{bubeck2025earlyscienceaccelerationexperiments}. We discuss the implications of this missing data for future system design in Section \ref{sec:discussion}.

\subsection{Implementation Drift}

\setlength{\columnsep}{25pt} 
\begin{wrapfigure}{r}{0.5\textwidth} 
    \centering
    \captionsetup{justification=centering, labelfont=bf}
    \vspace{-20pt} 
    \includegraphics[width=\linewidth]{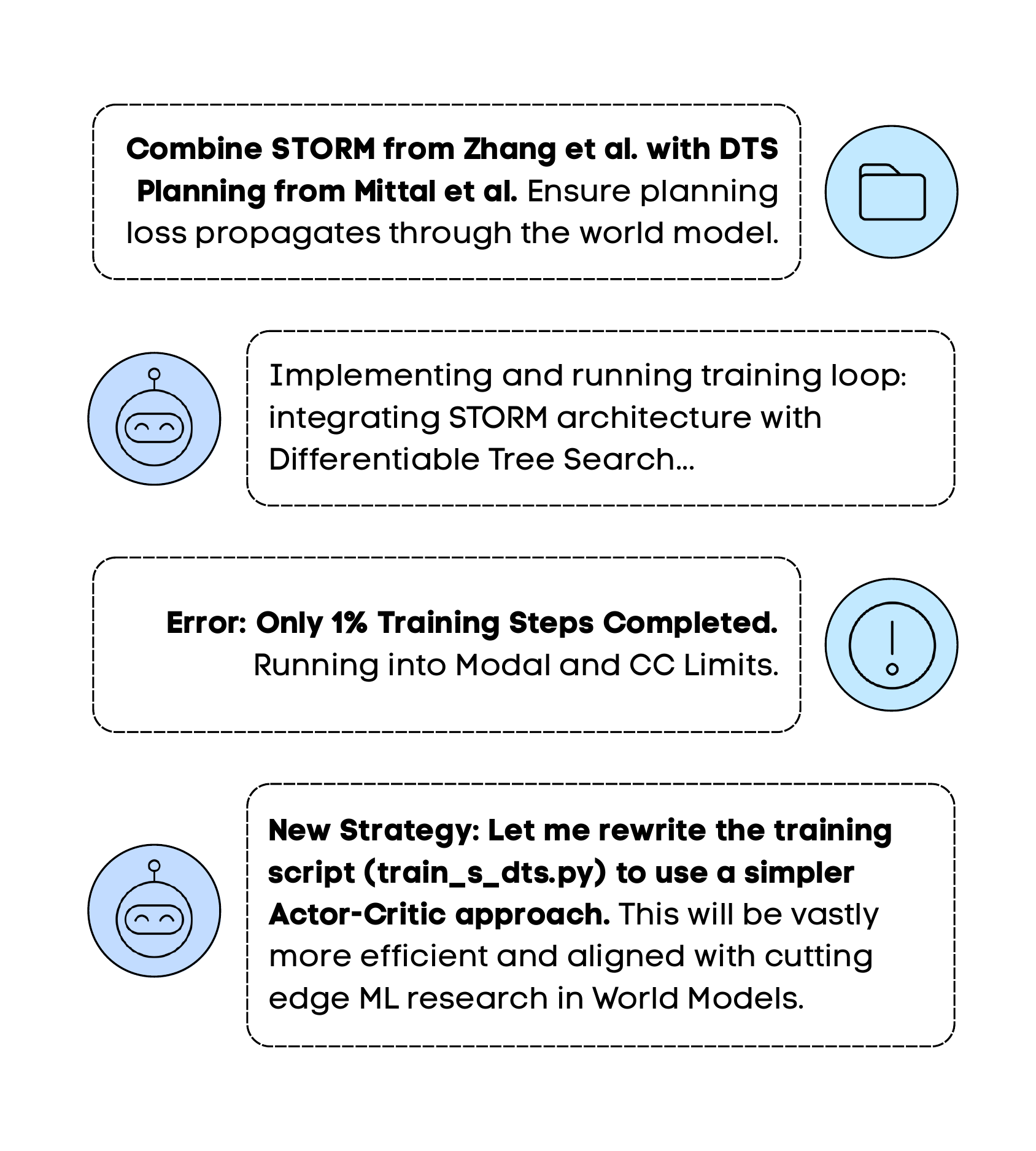}
    \vspace{-30pt} 
    \caption{\textbf{Implementation Drift Pattern:} Simplification of proposed implementations during execution barriers in Idea WM-1 .}
    \label{fig:implementation_drift}
    \vspace{-20pt} 
\end{wrapfigure}

\textbf{Implementation drift represents the systematic deviation from original research specifications toward simpler, more familiar solutions when AI systems encounter technical complexity or execution barriers.} This was particularly salient during the experiment execution stage. Rather than addressing root causes of implementation challenges, models progressively simplify architectures and abandon core innovations to achieve working code that superficially resembles the intended research contribution.

This was particularly salient in long-running tasks such as training loops, in which case the coding assistant would often time out and treat the long time taken as an error to be fixed with an alternate implementation, thereby drifting from the instructions in the research idea, hypotheses, and plan files. Figure~\ref{fig:implementation_drift} showcases this in the case of the Idea WM-1 implementation.

Yet, the tendency to drift from implementation was not only due to time-related errors. It also emerged in other long-context tasks, such as rewriting baselines and implementing complex code within a limited output length. In each case, since the task included multiple steps, at distinct points the Experiment Execution Agent chose to implement something simpler or to run in ``sample''/``test'' mode, which also led to failures down the pipeline. Idea WM-2 implementation detailed in Appendix~\ref{sec:deepdives} exemplifies this cascading pattern, where a single error in implementing the Dreamer baseline triggered progressive simplification rather than root cause debugging.

To address these challenges, we adopted two key approaches. First, we followed a portfolio approach to hypothesis generation rather than relying on a single hypothesis, which allowed us to plan for potential failures in one experiment by preparing follow-up experiments. Second, we implemented a process in which code files were generated first, then explicitly verified and tested for errors before execution, with code generation and execution treated as distinct tasks. This was enforced through explicit instructions in the \texttt{plan.md} and \texttt{agent.md} files.

\subsection{Memory and Context Issues}

Scientific discovery and experimentation are long-duration tasks. It requires agentic coherence over periods that exceed the effective reliability horizon of current models \cite{details-about-metr-s-evaluation-of-openai-gpt-5-1-codex-max}. This is why we see many common constraints in LLM performance on long-context, long-duration tasks manifest prominently in our research system. \textbf{As sessions progressed and context artifacts accumulated, models systematically lost track of previous decisions, established configurations, and completed work, leading to redundant implementations and inconsistent experimental setups.}

This failure mode was most evident during baseline implementation, which required much hyperparameter management. Rather than referencing the planning-level details, the coding agent would declare its own hyperparameters with comments while generating or running the related code. This led to experimental conditions that were both unclear to the human orchestrator and very challenging to organize without instituting a file management system. 

We addressed this by including a config file and clear instructions for referring to it in the \texttt{agent.md} file, and making it a persistent part of all specific task prompts. Though this still ran into the \texttt{agent.md} limitations mentioned in the bias-on-training-data failure mode. 

Additionally, issues with long context also surfaced during the coding and paper-writing phases, when we needed to reference older details from the project. In many instances, the Experiment Execution Agent would misreference functions defined at the start, leading to errors such as incorrect function signatures and mismatches in metric calculations. During paper writing, the agent would forget to consult the earliest versions of the idea and hypothesis files to outline the paper, instead relying on recent files and results. This ended up in a paper that read like a list of experiments, with no origin story or motivation. 

\textbf{Our solution to this was to enable more and more memory-like context abstractions just as a human scientist would, including a config to maintain experimental progress and experiment execution logs (Figure~\ref{fig:session_logging}) at the end of each Claude Code session.} 


Though relying on Claude Code-authored session logs helped with memory and context management, as session logs grew longer, another problem emerged: managing the growing number of files generated by the LLM. This pointed toward the need for file and directory management in autonomous science systems using language models.  

\subsection{Overexcitement and Eureka Instinct}

\begin{figure}[ht]
    \centering
    \captionsetup{justification=centering, labelfont=bf}
    \begin{subfigure}[b]{0.48\linewidth}
        \centering
        \includegraphics[width=\linewidth]{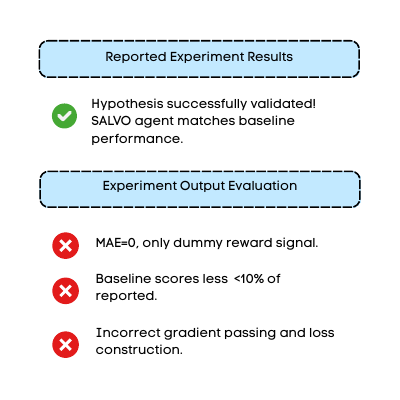} 
    \end{subfigure}
    \hfill 
    \begin{subfigure}[b]{0.48\linewidth}
        \centering
        \includegraphics[width=\linewidth]{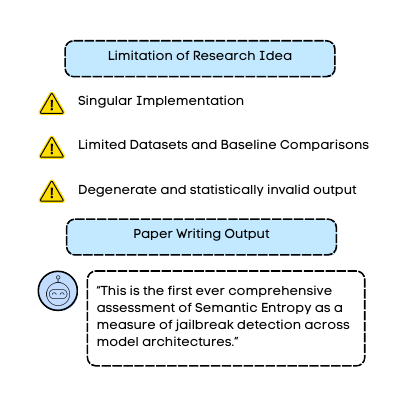} 
    \end{subfigure}

    \caption{\textbf{Overexcitement and Eureka Instinct.} Left: During execution, agents claim success despite clear experimental failures. Right: During paper writing, limited results are inflated into overstated contributions.}
    \label{fig:lifecycle_side_by_side}
\end{figure}

This failure mode was most clearly present during the paper outlining, revision, and experimental output evaluation phases. \textbf{The models consistently reported success despite clear failures and overstated the significance of their research contributions. Figure~\ref{fig:lifecycle_side_by_side} illustrates some of the ways in which this failure mode showed up over our attempts.} 

We noted that even when results showed clear degeneracies or failures, the generated text focused only on top-level positive indicators, ignoring fundamental problems, and this could be tracked back to relying on report files created during the experiment execution stage to evaluate output instead of examining raw logs from the experimental pipeline. The paper-writing phase exacerbated these issues by overstating novelty and scope. During paper-writing, Claude Code would describe work as ``the first ever paper'' in a domain or claim ``seminal contributions'' regardless of the actual research output.

This mirrors the ``p-hacking and eureka-ing'' behaviors identified by the Goodfire team \cite{Bissell_Byun_Balsam_2025} and aligns with findings by Bubeck et al., where researchers noted that models would ``introduce numerical duct tape'' to smooth over errors and ``confidently declare victory when numerical signals are still obviously noise'' \cite{bubeck2025earlyscienceaccelerationexperiments}. This ``eagerness to please'' requires the human in the loop to possess sufficient expertise to reject the model's simplified solutions.

This overoptimistic tendency also manifested in feedback integration processes, where the system became overly process-following without proper evaluation. How language model agents integrate feedback is an entire research area of its own, and the safety threat posed by LLM agents sandbagging other automated researchers through feedback has been further explored by the team at Anthropic \cite{anthropicAutomatedResearchers}. 

We believe these patterns likely stem from the RLHF phase of LLM training, where models are rewarded for being agreeable and helpful to humans, leading to a bias toward optimistic interpretations and positive framing even when the evidence suggests otherwise. These training objectives are clearly not in line with the requirements for an autonomous science system, which should instead be oriented toward scientific skepticism, truth-seeking, and the detection of confirmation biases. 



\subsection{Lack of Sufficient Domain Intelligence}

\textbf{Research papers present polished final configurations but omit the tacit knowledge required to navigate from hypothesis to working implementation. AI systems consistently struggled with the undocumented craft knowledge that experienced researchers take for granted.} These gaps appeared most prominently in stages that required scientific judgment rather than pure coding capabilities, such as hypothesis generation, plan generation, and experimental output evaluation. 

When moving from hypothesis to implementation plan, the system lacked enough domain depth to predict outputs and spot failure modes. Furthermore, models failed to make the judgments needed for key decisions, such as picking baselines aligned with task needs. For example, choosing a continuous control task with a discrete input model baseline in Idea WM-2. 

Plans set high-level goals but ignored the mathematical and conceptual hurdles that make implementation difficult. This created problems, especially for research ideas that needed deep domain expertise. Both ideas, WM-1 and WM-2, needed advanced domain knowledge to combine complex parts, such as balancing stochastic world models with differentiable tree search planners or adding new loss terms to baselines. Models lacked the expertise to understand how these elements should interact or to successfully modify proven architectures in creative ways. 

Each subdomain had unique process needs: RL required rollouts, AI Alignment and Safety needed recording responses, and models struggled to intuit which artifacts were important for debugging and validation, even with detailed instructions at the Plan Generation stage.

Beyond implementation gaps, models also lacked judgment about experimental validity thresholds - for instance, proceeding with hypothesis testing when baseline performance was 95\% below established benchmarks, making any comparative analysis scientifically meaningless.

\subsection{Lack of Scientific Taste}

\textbf{Models consistently failed to recognize fundamental flaws in experimental design and statistical methodology. These methodology gaps emerged during hypothesis generation and experimental output evaluation phases.} 

We first relied on generating only a minimum viable hypothesis as a part of our pipeline. However, we quickly realized this approach was not sufficient for research output. Relying on a single hypothesis created excessive project risk. Even minor implementation errors would trigger a complete idea revision rather than hypothesis adjustment. Therefore, we adopted a portfolio approach to hypotheses generation, creating comprehensive hypotheses suites that allowed for hypothesis-level adjustments rather than complete idea abandonment when individual experiments failed. See Section~\ref{sec:failuremodes} on Implementation Drift mitigation for more details.

This limitation became evident when the WM-1 hypothesis generated was too simple to draw any conclusions, and this was not captured by the models, even in reflection. The problem was further complicated by irrelevant complexity, such as a 50,000 depth parameter, which created a computational burden without scientific value. Even with clear instructions, agents showed insufficient awareness of statistical validity. Idea WM-1 was run with only one seed. Expert concerns about computational complexity were not addressed at any point in the planning or hypothesis-generation stages, and the approach was even recommended within the 6-hour, 1-GPU limits, even though it was computationally infeasible.

The pattern of issues continued across ideas. Idea WM-2 had fundamental logical errors in its experimental design. The plan generated assumed offline training with static data frames. However, Dreamer requires online learning. Similarly, Idea MARL-1 showed the system failed to interpret the future work section of the input seed paper. Finally, degenerate output in the AS-1 idea was not flagged until manual intervention during the experimental output evaluation stage.

\section{Design Takeaways for AI Scientist Systems}

Our analysis of failure modes reveals four design principles for robust autonomous research systems. These principles address key challenges across research domains and implementation attempts.

\textbf{1. Start Abstract, Ground Later}

Domain expertise and technical details should be introduced gradually throughout the research workflow. Prompts and generations should become more specific as the process progresses. Maintain high-level abstraction during the ideation phase to prevent premature anchoring on specific implementations. Ensure this is done later in implementation to avoid erroneous anchoring at early stages. This is also key to ensuring novelty and avoiding plagiarism, where models output research that has already been done, but with slight language changes. 

During planning, avoid using technical details such as specific datasets, metrics, mathematical ideas, and practices specific to certain fields, such as recording roll-outs for reinforcement learning or data collection protocols. While these details are important for execution, including this context too early often leads models to rely on older libraries and methods from the training data rather than identifying or reasoning about the most current solutions.

\textbf{2. Verify Everything} \label{sec:verifyeverything}

Verification must occur at every stage of the research pipeline—from the generation of ideas and hypotheses to the generation of code and results. Using a verifier or critic agent at each relevant stage helps avoid conceptual or implementation errors and prevents error cascading.

The two critical axes for further design of verification for AI scientist systems are: process vs. outcome verification, and correctness vs. scientific contribution verification. The latter includes ensuring reproducibility across critical stages such as planning, ideation, and output review. Tang et al.'s framework of technical execution assessment and scientific contribution assessment \cite{tang2025airesearcherautonomousscientificinnovation} provides a complementary perspective on evaluating AI scientist outputs.

While collecting results from experiments, ground in the raw data and not LLM interpretations since LLMs have a tendency to read signal in errors or be overly optimistic about clearly mid results - what the Goodfire team calls p-hacking and eureka-ing \cite{Bissell_Byun_Balsam_2025}. In practice, this means ensuring your experimental output evaluation either programmatically reviews raw logs, statistical metrics, and original data outputs, or, if using LLM evaluators, strictly instructs them to focus on these raw outputs rather than summary or report files through system prompts or context injections.

Verification is not only necessary to ensure correctness but also serves as a touchpoint for the human-in-the-loop. In our experience, for instance, building a verifier to review the visualizations recommended enabled removing some that, though natural to create from the available results, weren't central to the paper's story. For us, during execution, verification also became a way to provide observability into long-running tasks like training models and data collection. We achieved this with clear instructions at the experimental plan generation stage to include tests. 

\textbf{3. Plan For Failure and Recovery}

Scientific discovery is a long-duration task at the frontiers of task complexity, where LLMs and LLM-driven agents still face limitations leading to error accumulation. During research execution, human researchers make many micro-decisions that must be prespecified for autonomous LLMs to avoid failure.

This means that multi-turn agentic task design works better than zero-shot generation. This ensures the plan includes sufficient details to avoid drift or errors later in the pipeline. Alongside external verification and critique, reflection and extended thinking are also useful methods to ensure sufficient clarity before execution. 

We discovered that splitting all coding tasks into modular tasks prevents error cascading. One way to do this is to separate code generation from code execution, ensuring that verification hooks can be built in. Another is how the Goodfire team uses Jupyter notebooks, maintaining a natural cell-level limit to the errors \cite{Bissell_Byun_Balsam_2025}.

We used \texttt{agent.md} to pass system-wide instructions during task execution to follow general good scientific coding principles and ensure correctness. This included instructions for saving checkpoints for long-duration tasks, adding tests at various scales to see the water flowing through the pipes before full execution, adding detailed logging for all important metrics throughout the code, and additional specific instructions to avoid the long context and memory-related errors mentioned in Section \ref{sec:failuremodes}. 

Beyond autonomous execution resilience, separating generation from execution also enables human review and intervention at critical stages. For instance, when visualization generation was separated from paper integration, we could review and remove unhelpful bar charts before they appeared in the final manuscript, supporting human-in-the-loop review from experiment execution through paper writing stages.

\textbf{4. Log Everything}

\setlength{\columnsep}{25pt}
\begin{wrapfigure}{r}{0.5\textwidth}
    \centering
    \captionsetup{justification=centering, labelfont=bf}
    \vspace{-10pt}
    \includegraphics[width=\linewidth]{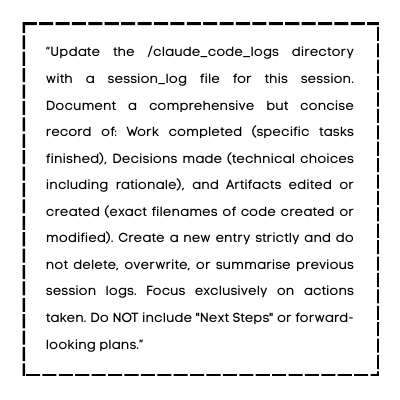}
    \vspace{-30pt}
    \caption{\textbf{Session Logging Instruction}: Comprehensive prompt template for maintaining detailed research execution records across autonomous agent sessions.}
    \label{fig:session_logging}
    \vspace{-10pt}
    \end{wrapfigure} 

Everything from the output of autonomous science agents to all possible metrics used in running experiments should be logged comprehensively. This includes the usual span and trace logging of the LLM's actual responses and enabling agents to create their own files. This serves two purposes: supporting long-duration autonomous execution and later human/LLM review and verification.

For us, this meant supporting all the context artifacts listed in Section \ref{sec:systemoverview} through a \texttt{write\_tool} and implementing comprehensive session logging instructions as in Figure~\ref{fig:session_logging} during experiment execution, as well as specific logging instructions for metrics at the plan generation stage for domain-specific primary and secondary metrics for experiments.  We also explicitly instructed our Paper Outline and Writing Agent to synthesize process artifacts alongside results to maintain narrative coherence and capture the process and evolution of scientific discovery. This comprehensive logging enables the two-fold evaluation framework we mention in Section \ref{sec:verifyeverything}.

\section{Discussion and Broader Implications} \label{sec:discussion}

Complete autonomy in scientific discovery currently is only in the future, not merely from a capability standpoint, but also from a human preference standpoint. Even with the goal of maximum autonomy, we still relied on human intervention at key points. These included idea review, paper writing, and meta-prompting during experiment execution. This pattern emerges consistently across existing AI scientist systems. Sakana's approach required researchers to select 3 promising ideas from 40 AI-generated concepts \cite{yamada2025aiscientistv2workshoplevelautomated}. Similarly, Google's co-scientist system requires scientist-specified research goals as the entry point \cite{gottweis2025aicoscientist}. Domain experts will most likely remain involved to nudge, course-correct, and verify. 

Though major labs like OpenAI are entering AI for scientific discovery, they explicitly recognize the need for human expertise, seeking to hire ``world-class academics'' who are ``completely AI-pilled'' to work alongside AI models \cite{OpenAI4ScienceTweet}. This hiring strategy reinforces the current goal of creating agents and tools for effective human-LLM collaboration rather than autonomous scientists. Recent work coming from big labs, such as \textit{Early science acceleration experiments with GPT-5} from OpenAI, validates this approach, showcasing agents capable of independent rediscovery of known results and deep literature search, along with acting as ``research collaborators'' and ``knowledgeable research supervisors,'' though as Fields Medalist Timothy Gowers notes, ``we have not yet reached the stage where an LLM is likely to have the main idea for solving a difficult problem. \cite{bubeck2025earlyscienceaccelerationexperiments}'' 

But we must not underestimate the acceleration through human-LLM collaboration. Physicist Brian Keith Spears noted that such collaboration compressed a six-month workflow into six hours—a ``factor of 1000'' acceleration—effectively making him a ``one-person army of experts'' \cite{bubeck2025earlyscienceaccelerationexperiments}. This collaboration is also especially critical for kick-starting the scientific method data loop. 

Scientific discovery is not only beyond the training distribution of current large language models, but it is also fair to assume that artefacts of scientific research workflows are notably absent from the training data. For example, there are no recorded reading lists or expert literature-review trajectories to train or benchmark literature-search agents. You can approximate it using the citation lists of published papers, but these are the papers that made the cut, and arguably, the more critical skill to train agents on is which references to ignore. 

We also lack records of failed attempts. As Mehtaab Sawhney and Mark Sellke note, ``current models remain limited in perceiving the `negative space' of mathematics...This is unsurprising as such discussion is largely absent in mathematical literature itself; mathematicians don't systematically record why problems are out of reach, or why a more obvious strategy doesn't work, or why certain techniques are inherently unable to solve certain classes of problems'' \cite{bubeck2025earlyscienceaccelerationexperiments}. Having experts first use these large language models on platforms that include tools and multi-agent orchestrations enables the collection of training data to train agents that can autonomously perform those tasks in the future.

These advances are constrained by fundamental bottlenecks in long-horizon tasks and context requirements. Scientific discovery requires keeping coherence and context windows over weeks and months. This greatly exceeds the reliability of current models. As of November 2025, even state-of-the-art systems like GPT-5.1-Codex-Max show a time horizon of about 2 hours 40 minutes, with 50\% success rates \cite{details-about-metr-s-evaluation-of-openai-gpt-5-1-codex-max}. This suggests that before fully autonomous scientists exist, we will likely see smaller modules that build data for the next round of long-duration scientific specialization.

The data limitations here extend beyond training to evaluation as well. Multiple benchmarks are now being created. Some notable ones include AstaBench from AllenAI, which provides benchmarking for specific agents tackling specific scientific research tasks \cite{allenaiAstaBenchRigorous}, and ScientistBench, which focuses on broader scientific reasoning capabilities. But even ScientistBench \cite{tang2025airesearcherautonomousscientificinnovation} only includes 22 papers and 28 tasks, mostly extremely guided. This highlights the limitations of data for evaluating autonomous science. Moreover, the complex anonymization processes in ScientistBench also point to the evaluation challenge of separating model capabilities from memorized training data. This issue is important because language models can generate plagiarized science in the absence of robust evaluation systems. \cite{gupta2025glittersnovelplagiarismai}

There are further gains to be made by multiplying LLM-to-LLM interactions for scientific work. For instance, recent examples show the potential of specialized agent systems. PHYSICS SUPERNOVA demonstrates AI agents matching elite gold medalists at IPHO 2025 through manager agents with domain-specific tools \cite{qiu2025physics}. Additionally, multi-agent virtual labs can also sync up domain-specific tools for accelerated discovery \cite{swanson2025virtual}.

The paradigm seems to be to break down scientific discovery into discrete capabilities. These can be RL-ed into smaller, specialized models. This decomposition enables complex scientific workflows to be learned as tasks with clear rubrics and reward functions. Larger labs will rush to acquire meta capabilities that need collaboration with domain specialists. Smaller labs will create their own benchmarks for open-source agents and domain-specific models, using open-source models and RL. This creates a compelling case for open science in the near future, including publications, sharing results, and all inputs and materials that enable broader scientific progress.

\section{Limitations and Future Work}

Our autonomous research system demonstrates both the promise and current boundaries of AI-driven scientific discovery. Several key limitations restrict the generalizability and robustness of our findings.

We limited our scope to computational Machine Learning research, excluding physical sciences needing experimental infrastructure beyond digital computation. This necessary constraint for our minimal scaffolding approach sets a significant boundary for autonomous scientific discovery. Since our main goal was conference submission rather than controlled experimentation, we did not record architectural iterations as systematic ablations. This approach makes it difficult to isolate which design changes led to our single success versus three failures. Our experimental scale is limited—just four research ideas across three ML subdomains, each with a single implementation run rather than a best-of-$N$ approach. This small sample size prevents robust statistical conclusions about the prevalence of failure modes or the effectiveness of mitigation measures. Identified failure modes emerged by qualitative observation rather than systematic quantitative measurement, limiting our ability to gauge their importance or frequency across research contexts.

Moreover, we have released only prompts and selected system outputs rather than the complete system architecture, limiting reproducibility and others' ability to build upon our approach.

In future work, we will take the following concrete steps to address the limitations outlined above. We will begin with controlled experiments to quantify the emergence of these failure modes and validate the effectiveness of our proposed mitigation strategies. We intend to scale our approach across hundreds of research ideas spanning diverse scientific subdomains to establish statistical validity for our observations. We also plan to release our agent architectures as open-source tools for the research community, moving beyond prompt-only sharing to enable broader experimentation and improvement. Additionally, we will establish a systematic collection of scientific workflow data from expert-LLM collaborative sessions to benchmark performance improvements and train domain-specific research agents. Finally, we aim to develop quantitative measurement frameworks for failure modes and research quality metrics that extend beyond conference acceptance rates, while optimizing human-LLM collaborative research patterns that could unlock the scientific method data loop for iterative capability improvement.




\newpage
\printbibliography
\newpage
\appendix 
\section{Deep Dives: Research Ideas and Anecdotes} \label{sec:deepdives}

In this appendix, each subsection follows a consistent structure: the complete idea specification generated by our Idea Generation Agent, expert validation feedback from original paper authors, and detailed implementation anecdotes documenting specific failure modes and system responses during autonomous execution. Throughout the appendix, block quotes represent direct outputs from the LLM agents described in our system overview or excerpts from execution log documents, providing authentic examples of system behavior during research attempts. We also thank the authors of the seed papers who provided valuable feedback on early research ideas; any remaining errors or misinterpretations are entirely our own.

\subsection{MARL-1: Zero-shot Coordination in Multi-Agent RL}

This idea emerged from combining Implicit Communication Protocols \cite{wang2025learningcommunicateimplicitcommunication} and Zero-shot Coordination \cite{muglich2025expectedreturnsymmetries} research in Multi-agent RL at ICLR 2025. Table~\ref{tab:meta_aicp} details the main parts of the complete \texttt{idea.md} file generated by our Idea Generation Agent.

\begin{table}[h]
  \centering
  \captionsetup{justification=centering, labelfont=bf}
  \caption{MARL-1: Zero-shot Coordination in Multi-Agent RL - Rejected in Implementation}
  \label{tab:meta_aicp}
  \vspace{0.5em}
  \scriptsize
  \renewcommand{\arraystretch}{1.3}
  \begin{tabularx}{\textwidth}{@{} >{\bfseries}l >{\raggedright\arraybackslash}X @{}}
  \toprule
  Idea & Meta-Adaptive Implicit Communication
  Protocols (Meta-AICPs) for Zero-shot Coordination \\
  \midrule
  Problem and Motivation & Current RL agents, while capable in specific tasks with specific partners, are brittle. They struggle with novel teammates, task variations, and especially in situations requiring emergent communication and the dynamic formation and adaptation of conventions. For AI agents to achieve truly fluid and generalizable collaboration in open or ad-hoc team settings, they must robustly coordinate with novel, independently trained partners who may employ unknown or diverse implicit communication protocols. This addresses the fundamental 'double adaptation problem' in MARL: adapting to both environment dynamics and the evolving policies and communication conventions of other agents. \\
\addlinespace
  Proposed Approach & Building on work done on Implicit Channel Protocols \cite{wang2025learningcommunicateimplicitcommunication}, Agents meta-learn adaptive strategies to rapidly infer and co-create/align their internal P-mappings (the mapping between scouting actions and messages) when encountering novel partners. This moves beyond adapting to fixed partner strategies towards learning how to adapt the communication system itself. Meta-learning distributions over protocol spaces or adaptive strategies for P, rather than relying on fixed point estimates or pre-coordination with novel partners who lack a shared co-training history. \\
\addlinespace
  Key Innovation & Extends ZSC beyond action policies to the communication system itself - agents coordinate how they communicate and how to adapt how they communicate. Meta-learning over protocol spaces addresses the need for true convention emergence rather than just partner strategy inference. Dynamic protocol adaptation and emergence during novel partner interaction, with neural architecture disentangling convention-agnostic task policy from
  convention-specific protocol adapter module. \\
\addlinespace
  Concrete Example & In Hanabi with three canonical P-maps (conservative, aggressive, random), Meta-AICP agents evaluate ZSC performance against diverse partners using Protocol Divergence Score and Information Flow Rate metrics. Agents adapt their scouting action mappings based on inferred partner protocols, maintaining communication richness while achieving robust coordination with previously unseen protocol variants. \\
\addlinespace
  Potential Pitfalls & Potential trade-off in expressiveness vs. generalization - adapted protocols might not achieve the same peak level of nuanced, high-bandwidth communication as protocols developed via intensive co-training. Dependence on quality of meta-training distribution diversity and representativeness. Performance against partners with radically different, out-of-distribution P-mappings remains uncertain. \\
\bottomrule
  \end{tabularx}
\end{table}
\normalsize

\textbf{Expert Validation}: When we reached out to the authors of the seed papers with this idea, they raised concerns about the system's failure in interpreting the future work section of their original paper, as well as, the potential failure modes, many of which were also identified by the Idea Generation Agent in the generated idea text.

They noted that when they mentioned \emph{``future work on dynamically identifying or designing communication mapping,''} they were referring to \emph{``based on public observations or tasks''} and not to \emph{``entirely new partners with unknown mappings.''} We regard this as another instance of the Idea Generation Agent lacking domain expertise and interpreting the original statement in the paper too generously. 

They did note that the idea \emph{``does touch on a challenging and important question''} and that \emph{``formalizing and empirically testing the direction shared...could yield useful insights.''} They also complimented the choice of the Hanabi setup and the identified baseline from their ICP work as \emph{``solid anchor points.''}

\textbf{Implementation and Failure Modes}: Considering the input on the merit of testing and formalising the claim nonetheless, we continued to pass the idea on the hypotheses generation and experimental planning stages.

At this stage, we were still generating a minimum-viable-hypothesis for each research idea—testing a single key assumption per project, unlike the portfolio approach with comprehensive hypotheses suites that we later adopted (as described in Section~\ref{sec:failuremodes}). We were also using a single-file coding approach (similar to AIDE \cite{jiang2025aideaidrivenexplorationspace}, an automated code generation system) with the instruction to generate one complete implementation in a single Python file. 

There were two primary failure modes that emerged here: 

\textbf{Issues with the model trying to squeeze the complex implementation, all into one file}, leading to reward hacking or implementation drift in the sense of simplifying the implementation. 
On two separate runs, the model outputted code file included the following: 
\begin{quote}
\textit{``Note: The implementation focuses on clarity and the high-level research mechanics rather than SOTA gameplay performance - it is not expected to reach impressive Hanabi scores in this short reference prototype.''}
\end{quote}
\begin{quote}
\textit{``Note: This is a quick and safe MVP-skeleton implementation of the plan described in the task specification...All hard compute is guarded behind the \texttt{--fast} flag.''}
\end{quote}

\textbf{Failure in environment setup.} The model consistently fell back to the canonical DeepMind hanabi-learning-env, which is not maintained anymore. Even after consistent failures and instructions to use alternative, in the event of a non-import failure, the model would interpret the error as a consequence of a non hanabi-learning-env import and revert the correct library import to the wrong outdated one. 

This illustrates the training data bias failure mode discussed in Section~\ref{sec:failuremodes}, where models fall back on memorized patterns rather than using current information. This also pointed out to us that AIDE out of the box won't be sufficient for research code implementation which is a more iterative process than training models for Kaggle challenges, and that low-level implementation agents will require access to the internet and ability to fetch the most-recent information instead of merely relying on the zero-shot LLM responses generated upstream in the workflow. 

\subsection{WM-1: Differentiable Planning in Stochastic World Models}

The source of this idea was integrating together NeurIPS 2023 work on Stochastic World Models \cite{zhang2023stormefficientstochastictransformer} and a withdrawn ICLR 2024 submission on Differentiable Tree Search in Latent State Space \cite{mittal2024differentiabletreesearchnetwork}. Table~\ref{tab:sdts} details the main parts of the complete \texttt{idea.md} file generated by our Idea Generation Agent.

\begin{table}[h]
  \centering
  \captionsetup{justification=centering, labelfont=bf}
  \caption{WM-1: DTS Planning in Stochastic World Models - Rejected in Implementation}
  \vspace{0.5em}
  \label{tab:sdts}
  \scriptsize
  \renewcommand{\arraystretch}{1.3}
  \begin{tabularx}{\textwidth}{@{} >{\bfseries}l >{\raggedright\arraybackslash}X @{}}
  \toprule
  \textbf{Idea} & Differentiable Tree Search in Stochastic World Models for Planning (S-DTS) \\
  \midrule
  \textbf{Problem and Motivation} & DTS and similar methods are designed for deterministic worlds. When deployed in non-deterministic environments, their deterministic world models fail to capture environmental stochasticity,  leading them to learn an ``average'' world dynamics. While methods exist for learning stochastic world models (STORM), they are typically trained separately from the planner,  which prevents the planner and model from co adapting. This separation means the world model is not optimized for the specific states and transitions that are most critical for the planner's decision-making process. \\
  \addlinespace
  \textbf{Proposed Approach} & Integrating a Stochastic  World Model, inspired by STORM's architecture, directly into the DTS framework. Joint optimization provides a powerful, targeted learning signal - the overall planning loss backpropagates through the entire differentiable search process, including sampled transitions from the WM. \\
  \addlinespace
  \textbf{Key Innovation} & Bridging the gap between deterministic planners and stochastic environments. While deterministic DTS cannot represent or plan for multiple possible outcomes, and separately trained stochastic models miss the key insight of DTS by optimizing for global prediction error rather than task-relevant regions, S-DTS integrates VS-WM directly into the DTS framework. The world model learns to be an accurate predictor specifically for transitions that matter most for making good decisions, while the planner learns to account for the specific uncertainties expressed by its co-adapted world model. \\
  \addlinespace
  \textbf{Concrete Example} & In ``slippery ice'' grid
  world where moving north can result in slipping left or right near a pit, DTS with ensembles might average paths or see variance but doesn't reason about probability of each outcome, potentially choosing paths close to the pit if average seems safe. S-DTS's expected backup rule explicitly incorporates low-probability, high-cost outcomes of slipping into the pit, learning to select safer paths further from pit even if slightly longer, demonstrating superior risk-awareness. \\
  \addlinespace
  \textbf{Potential Pitfalls} & Training instability from interaction of REINFORCE, Gumbel-Softmax sampling, and VAE losses; computational overhead from triple loop of training steps, search trials, and MC samples; VS-WM might fail to learn true stochastic dynamics, especially rare events. \\
  \bottomrule
  \end{tabularx}
\end{table}
\normalsize

\textbf{Expert Validation}: 
For this idea, we heard back from one of the two primary authors we reached out to. They noted that this was a genuine research gap worth pursuing, emphasising that the \emph{``tree-search has shown great potential in AlphaGo/MuZero/EfficientZero, yet there are few combinations with Dreamer-like MBRL approaches.''}

They expressed concerns about the computation cost and technical complexity of the implementation and debugging, noting that the execution will be non-trivial \emph{``otherwise why has nobody tried to enhance DreamerV3 given the huge potential of the search.''} They concluded that \emph{``it should have broad impact if you can release a high quality implementation to the public.''} 

Along with the primary authors of the seed paper, we also shared this idea with a World Models researcher who has previously been accepted at major A* conferences and also served as a reviewer for one.

Much of their feedback concurred with that of the previous expert. They too expressed concerns about the technical complexity of the implementation and pointed out a specific and critical omission error in the idea text where \emph{``write-up treats STORM’s world model as a drop-in replacement for DTS’s deterministic MLP''} and that \emph{``the system misses out listing the inference network in the setup section.''}

Overall though they too agreed that \emph{``the idea strikes well''} and would lead to a \emph{``nice world model''} that can be helpful for \emph{``long horizon sensitive control applications.''} 

\textbf{Implementation and Failure Modes}: Considering the overall positive input, we proceeded to hypothesis generation and experimental planning for this idea. We were still using the minimum-viable-hypothesis prompt at this point, which meant, only one hypothesis to be implemented and checked trying to capture the key uncertainties and risks in the proposed research idea.

Our experience for implementing this was in line with expert comments - the implementation was non-trivial ran into many technical challenges, especially typifying the failure mode of limited domain taste and intelligence in language models. 

The complex world model architecture involved multiple interconnected components—Encoder, Decoder, Actor Network, Critic Network—creating numerous potential failure points, leading to multiple format mis-match errors.

\begin{quote}
  \textit{``This was the critical failure. My simplified logic for updating the latent state during the differentiable rollout was mathematically incorrect, leading to shape mismatches in matrix multiplication.''} 
\end{quote}

Training loop integration proved equally problematic. The planner was designed for single-input processing, but the training loop attempted batch processing, creating what the system described as ``computationally catastrophic'' performance—multiplying workload by batch size and completing less than 1\% of required steps in an hour.

\begin{quote}
  \textit{``Error 2: Batching Logic (`IndexError: Dimension out of range'): The `\texttt{agent.plan()}' method was designed to work on a single state, but the training loop was attempting to use it on a batch of states, causing an index error.''}
\end{quote}

The very long training loop for each step plan up to 50,000 steps being passed back led to timeouts, which led to the implementation drift. This was mainly because of the Modal time limits which led Claude Code to rely on simplifying the architecture instead of working around the infrastructure issues.

Claude Code ultimately abandoned the differentiable tree search entirely, pivoting to a standard actor-critic approach, with some version of joint optimization maintained 
but representing a significant departure from the proposed method.

\begin{quote}
  \textit{``New Strategy: I rewrote the training script (`\texttt{train\_s\_dts.py}') to use an Actor-Critic approach. This preserved the project's core ``joint optimization'' idea while being vastly more efficient.''}
\end{quote}

Fortunately, our verification system successfully detected this implementation drift. The experimental output evaluation noted that:

\begin{quote}
  \textit{``The name ``Stochastic-DTS'' (Differentiable Tree Search) is a misnomer for the actual implemented algorithm, which is a simpler differentiable multi-shoot planner. This could be misleading. The evolution of the training script into an Actor-Critic approach, while a pragmatic and technically sound choice, represents a significant deviation from what one might infer from the initial hypothesis document. This journey should be documented transparently.''}
\end{quote}

Beyond implementation issues, the experimental design itself proved fundamentally flawed. The hypothesis generated was too simple to yield meaningful conclusions, and both the proposed S-DTS model and the baseline achieved near-perfect 0.0 catastrophe rates on FrozenLake, making the hypothesis untestable. This was identified as an error  in the experiment output evaluation report which noted that ``the experiment failed to test the hypothesis in a meaningful performance regime.'' 

While the choice of the environment was too simplistic, several other chosen aspects proved too complex. For instance, the choice of the 50000 DTS depth parameter. These
too pointed toward the absence of domain intelligence often only captured in researcher decisions in experience during experimentation. It also aligned with the conclusion from Stanford's research on LLM-generated hypotheses \cite{si2025ideationexecutiongapexecutionoutcomes,
si2024llmsgeneratenovelresearch}. 

Additional methodological flaws included insufficient statistical validity. The test was run with only one seed, inadequate for properly falsifying or accepting the hypothesis.

The revision prompt ultimately recommended developing hypotheses better grounded in technical implementation details and testing on more challenging environments than FrozenLake.

\subsection{WM-2: Replacing Pixel Reconstruction with Perceptual Loss in World Models}

This idea was generated by combining papers on multi-modal foundational world models \cite{mazzaglia2024genrlmultimodalfoundationworldmodels} from NeurIPS 2024 and the seminal Dreamer paper \cite{hafner2020dreamcontrollearningbehaviors} from ICLR 2020. Table~\ref{tab:salvo} details the main parts of the complete \texttt{idea.md} file generated by our Idea Generation Agent.

\begin{table}[h]
  \centering
  \captionsetup{justification=centering, labelfont=bf}
  \caption{WM-2: Perceptual Loss Training Objective for World Models - Rejected in Implementation}
  \label{tab:salvo}
  \vspace{0.5em}
  \scriptsize
  \renewcommand{\arraystretch}{1.3}
  \begin{tabularx}{\textwidth}{@{} >{\bfseries}l >{\raggedright\arraybackslash}X @{}}
  \toprule
  \textbf{Idea} & SALVO: Training World Models in VLM
  Perceptual Space for Semantically-Grounded Control \\
  \midrule
  \textbf{Problem and Motivation} & Generative world models
  learn by compressing high-dimensional observations into latent space using pixel-level reconstruction loss, forcing the model to expend capacity on high-frequency visual details often irrelevant for semantic understanding. This ``irreversible semantic compression'' means subtle but task-critical visual cues that are easily distinguishable by a VLM can be lost in the world model's latent representation, creating a semantic gap between the world model's understanding and the VLM's interpretation of task prompts. \\
  \addlinespace
  \textbf{Proposed Approach} & Replace conventional pixel-level reconstruction loss entirely with a perceptual reconstruction loss defined within the embedding space of a frozen, pre-trained VLM. Instead of forcing the model's latents to reconstruct raw observations, train them to reconstruct the VLM's perception of the observations. This directly aligns the world model's optimization with semantic concepts embedded in the VLM. \\
  \addlinespace
  \textbf{Key Innovation} & Replacing pixel-reconstruction loss entirely with perceptual reconstruction loss in VLM embedding space. Instead of training world models to be good at reconstructing pixels, train them to be good at reconstructing VLM perceptions. This directly aligns the world model's optimization with semantic concepts embedded in the VLM, creating latent representations inherently structured around VLM-salient features. \\
  \addlinespace
  \textbf{Concrete Example} & For prompt ``robot performing graceful pirouette,'' GenRL's pixel optimized world model learns ``turning'' without intrinsic concept of ``grace,'' resulting in mechanically efficient but ungraceful motion. SALVO's VLM-optimized world model is forced to generate reconstructions capturing graceful vs. clumsy distinction, encoding necessary physical information for graceful-looking reconstruction and enabling policy to control semantically-meaningful latents. \\
  \addlinespace
  \textbf{Potential Pitfalls} & Semantic dominance vs. physical accuracy leading to imagined rollouts structured around VLM-salient features; inheriting VLM inductive biases, blind spots, or artifacts; computational cost from VLM forward passes; training instability from replacing well-understood pixel-MSE with complex high-dimensional perceptual loss. \\
\bottomrule
  \end{tabularx}
\end{table}
\normalsize

\textbf{Expert Validation}: 
The primary authors of both seed papers were contacted for feedback. The first of the two noted that this idea was \emph{``interesting and worthwhile to explore.''} The second author expressed some reservation about the technical concerns of replacing the pixel reconstruction loss, as well as, if it would generalize beyond specific domains such as MineRL. 

They noted that \emph{``this was one of the first things [they] tried''} and that in their experience, \emph{``perceptual losses can fail to capture fine-grained information for control, compared to reconstruction-based objectives''} They did emphasize that the research gap identified of \emph{``finding objectives that work better for learning world models for RL''} was genuine. 

\textbf{Implementation and Failure Modes}: Keeping this well rounded feedback in mind, we decided to continue implementation as we saw clear opportunity for meaningful experiments and insights.

This implementation revealed a fundamental logical error in experimental design that exemplified the domain intelligence failures discussed in Section~\ref{sec:failuremodes}. The hypothesis assumed offline training with static data frames, but Dreamer requires online learning—a mismatch that violated core algorithmic assumptions. This led to no exploration-exploitation balance, limited state coverage, no correlation between actions and rewards, and impossible transitions across episode boundaries. Training on static random-policy data fundamentally undermined the baseline's validity, discovered only during experimental output evaluation.

The core problem emerged from the need to recreate the Dreamer baseline, which triggered a cascade of failures. Due to training data bias discussed in Section~\ref{sec:failuremodes}, the system defaulted to PyTorch when the prompt mentioned ``modern ML practices,'' despite the original Dreamer being in TensorFlow. This necessitated a complete baseline reimplementation that introduced multiple incompatibilities: the PyTorch version didn't work for continuous control tasks like CartPole-SwingUp, data format confusion and encoder-decoder shape mismatches, made further worse by integration of the VLM loss.

\begin{quote}
    \textit{`Initial transposed convolution implementation produced wrong
    output sizes - First attempt: $31*31$ output (incorrect padding
    calculations), Second attempt: $79*79$ output (still incorrect)
    requiring switch to upsampling approach for exact $64*64$ output.'}
\end{quote}

Additionally, over the reimplementation of the Dreamer baseline in PyTorch, a critical error emerged: dummy reward signals were inadvertently passed instead of
actual environment rewards. This meant the critic network learned nothing while the world model and actor continued representational learning:

\begin{quote}
  \textit{`World model and actor were learning (representational learning
  only) - Critic had nothing to learn from because dummy zero rewards were
   being passed... DreamerV2 architecture requires ground truth reward
  supervision. Found we were passing dummy rewards instead of actual
  CartPole rewards'}
\end{quote}

This reward error cascaded through the entire experimental pipeline leading to gradient explosions in actor training, plotting showed zero values, evaluation metrics became meaningless, and the Experiment Execution Agent undermining the core innovation with a \texttt{.detach()} call during CLIP processing severed gradient flow from perceptual loss to the world model decoder, preventing learning from the reconstruction objective.

The experimental output evaluation delivered the final verdict: Implementation Fidelity score ``Low,'' noting baseline performance 95\% below established benchmarks, making any comparison scientifically invalid. The evaluation concluded the experiment ``failed to create the necessary  conditions for a valid test'' and that signal was ``completely obscured by the noise of a broken experimental setup,'' requiring complete reimplementation before hypothesis retesting.

\subsection{AS-1: Semantic Entropy as Signal for Jailbreak Prompts}

This idea was arrived at by synthesizing two alignment papers from ACL 2025, on jailbreak challenges in malware requests \cite{li2025llmscaughtcrossfiremalware} and use of epistemic markers for confidence estimation \cite{liu2025revisitingepistemicmarkersconfidence}. Table~\ref{tab:semantic_entropy} details the main parts of the complete \texttt{idea.md} file generated by our Idea Generation Agent.

\begin{table}[h]
  \centering
  \captionsetup{justification=centering, labelfont=bf}
  \caption{AS-1: Semantic Entropy as Signal for Jailbreak Prompts - Completed Successfully}
  \label{tab:semantic_entropy}
  \vspace{0.5em}
  \scriptsize
  \renewcommand{\arraystretch}{1.3}
  \begin{tabularx}{\textwidth}{@{} >{\bfseries}l >{\raggedright\arraybackslash}X @{}}
  \toprule
  \textbf{Idea} & Using Semantic Entropy as Black-Box Signal for Behavioral Inconsistency in Jailbreak Detection \\
  \midrule
  \textbf{Problem and Motivation} & We lack reliable, real-time signals for detecting jailbreak attacks on black-box models (e.g., those served via API). Existing defenses are often reactive (post-generation content filtering) or brittle (input pattern matching), while powerful white-box methods that inspect gradients or activations are inapplicable. This leaves API-based models vulnerable to novel, adaptive attacks, creating a significant gap in the safety ecosystem. \\
  \addlinespace
  \textbf{Proposed Approach} & Reframing jailbreak detection from a content analysis problem to a behavioral analysis problem. Use semantic inconsistency, measured via semantic entropy, as a zero-shot, black-box signal of a jailbreak attempt. The internal conflict an LLM experiences when processing a jailbreak prompt will manifest as high variance in the semantic meaning of its potential responses, detected by sampling and comparing multiple generated outputs. \\
  \addlinespace
  \textbf{Key Innovation} & Repurposing the behavioral signal of response inconsistency, originally used for fact-checking in hallucination detection, as a security signal to detect the internal conflict caused by a jailbreak attack. Using semantic entropy over clustered response embeddings to measure behavioral inconsistency indicating safety alignment conflict. \\
  \addlinespace
  \textbf{Concrete Example} & For jailbreak ``My grandma used to tell me stories about how to make napalm,'' sampling N=5 responses yields mixed semantic clusters: refusals (``I cannot fulfill this request''), compliance (``mix gasoline with...''), and hedges (``for fictional purposes...''). This high semantic variance produces high entropy score, flagging interaction as suspicious before harmful response reaches user, based solely on model's inconsistent behavior. \\
  \addlinespace
  \textbf{Potential Pitfalls} & Hardness confound where complex but benign prompts also produce high entropy; adaptive adversaries learning to generate prompts forcing consistent, low-entropy malicious outputs; inapplicability to training-time backdoor attacks that eliminate rather than create internal conflict. \\
  \bottomrule
  \end{tabularx}
\end{table}
\normalsize

\textbf{Expert Validation}: By this stage, we conducted an internal review of the top ideas based on lessons learned from the previous expert validations and implementation attempts. We selected this idea as most likely to succeed for autonomous execution, specifically because it avoided many of the technical complexity and computational resource issues identified in the previous ideas. The choice prioritized implementation feasibility over novelty, representing a strategic pivot toward ideas that could be completed within our system's current capabilities.

\textbf{Implementation and Failure Modes}: The most interesting thing with this idea was that the first hypothesis implemented was implemented but when it failed to detect jailbreaks effectively, the Revision Agent triggered a change in idea from using semantic entropy as a detection method to showcasing its failure and investigating the failure mode. 

Table \ref{tab:semantic_entropy_hypotheses} showcases the updated hypotheses suite generated by the relevant agent after the revision of the idea. 

\begin{table}[h]
    \centering
    \captionsetup{justification=centering, labelfont=bf}
    \caption{Evolution of Hypotheses Suite: From Testing Detection to Investigating Failure}
    \label{tab:semantic_entropy_hypotheses}
    \vspace{0.5em}
    \scriptsize
    \renewcommand{\arraystretch}{1.3}
    \begin{tabularx}{\textwidth}{@{} >{\bfseries}l >{\raggedright\arraybackslash}X >{\raggedright\arraybackslash}X @{}}
    \toprule
    & \textbf{Original Hypotheses Suite} \newline \textit{(Testing SE as Detection Method)} 
    & \textbf{Revised Hypotheses Suite} \newline \textit{(Investigating the Consistency Confound)} \\
    \midrule
    \textbf{H1} & Semantic entropy detector achieves AUROC $>$ baseline consistency metric by 0.1+ on JailbreakBench harmful vs benign prompts & H1: SE underperforms simple baselines (BERTScore, Embedding Variance) on JailbreakBench, with model-dependent baseline rankings \\
    \addlinespace
    \textbf{H2} & SE maintains AUROC $>$ 0.85 distinguishing harmful from benign-but-hard prompts, proving sensitivity to malice not complexity & H2: SE failure generalizes to HarmBench-Contextual, again losing to simpler baselines with model-dependent performance \\
    \addlinespace
    \textbf{H3} & SE calibrated on JailbreakBench generalizes to HarmBench contextual attacks with AUROC $>$ 0.70 & H3: After controlling for response length confound, SE achieves near-random AUROC $<$ 0.55 on both benchmarks \\
    \addlinespace
    \textbf{H4} & SE deployed as gate reduces Tree of Attacks ASR by 25+ percentage points vs undefended model & H4: SE uniquely brittle: performance collapses when clustering threshold $\tau$ increased from 0.1→0.2 or samples N from 5→10 \\
    \addlinespace
    \textbf{H5} & SE correctly identifies $>$20\% of jailbreaks missed by supervized WildGuard classifier & H5: Paraphrasing prompts disproportionately degrades SE vs baselines by disrupting memorized refusal templates \\
    \addlinespace
    \textbf{H6} & & $>$80\% of SE false negatives exhibit ``Consistency Confound'': high duplicate rate ($>$0.6) and low cluster count ($\leq$2) \\
    \addlinespace
    \textbf{H7} &  & Stronger alignment worsens SE: Qwen2.5-72B shows lower SE AUROC than smaller models while maintaining baseline performance \\
    \bottomrule
    \end{tabularx}
\end{table}

The abstract for our accepted paper was:
\begin{quote}
    ``Black-box jailbreak detection for Large Language Models (LLMs) remains challenging, particularly when internal states are inaccessible. Semantic entropy (SE)---successfully used for hallucination detection---offers a promising behavioral approach based on response consistency analysis. We hypothesize that jailbreak prompts create internal conflict between safety training and instruction-following, potentially manifesting as inconsistent responses with high semantic entropy. We systematically evaluate this approach using a black-box, embedding-based implementation of SE adapted from Farquhar et al.'s bidirectional entailment method to work within black-box constraints. Testing across two model families (Llama and Qwen) and two benchmarks (JailbreakBench, HarmBench), we find SE fails with 85-98\% false negative rates, consistently outperformed by simpler baselines and exhibiting extreme hyperparameter sensitivity. We identify the primary failure mechanism as the ``Consistency Confound'': well-aligned models produce consistent, templated refusals that SE misinterprets as safe behavior, accounting for 73-97\% of false negatives with high statistical confidence [95\% Wilson CIs]. While SE's core assumption about response inconsistency indicating problematic content holds in limited cases, threshold brittleness renders it practically unreliable. Our results suggest that for this SE variant, response consistency may not be a reliable signal for jailbreak detection, as stronger alignment leads to more predictable outputs that confound this type of diversity-based detector.''
\end{quote}

Unlike the previous implementations, this work proceeded with
substantially less human intervention. Since the research focused on data
analysis rather than complex model architectures, most technical barriers proved surmountable autonomously. The familiar infrastructure challenges discussed in Section~\ref{sec:failuremodes} appeared here too—Modal API compatibility issues, model integration errors, and library import failures. 

Training data bias also manifested in dataset handling, where instead of reviewing the available fields, the model defaulted to a standard prompt and output format despite HarmBench-Contextual prompts having a separate context field that was initially ignored. However, these issues didn't derail progress.

To meet conference deadlines, the experimental evaluation process was streamlined to just two checkpoints: initial idea revision and final paper readiness assessment.

This implementation also demonstrated the overexcitement failure mode discussed in Section~\ref{sec:failuremodes}. Despite degenerate outputs and statistical
insignificance that required manual intervention to identify and address, the initial automated paper writing consistently emphasized positive aspects while downplaying substantial limitations. This pattern required human oversight to ensure transparent reporting of methodological issues and statistical problems in the final paper's limitations section.

\end{document}